# FPGA-QHAR: Throughput-Optimized for Quantized Two-Stream Human Action Recognition on The Edge

Azzam Alhussain
*College of Engineering and Computer Science*
*University of Central Florida*
Orlando, Florida, USA
azzam@ucf.edu

Mingjie Lin
*College of Engineering and Computer Science*
*University of Central Florida*
Orlando, Florida, USA
milin@ucf.edu

*Abstract*—Accelerating Human Action Recognition (HAR) efficiently for real-time surveillance and robotic systems on edge chips remains a challenging research field, given its high computational and memory requirements. This paper proposed an integrated end-to-end HAR scalable HW/SW accelerator co-design based on an enhanced 8-bit quantized Two-Stream SimpleNet-PyTorch CNN architecture. Our network accelerator was trained on UCF101 and UCF24 datasets and implemented on edge SoC-FPGA. Our development uses partially streaming dataflow architecture to achieve higher throughput versus network design and resource utilization trade-off. We also fused all convolutional, batch-norm, and ReLU operations into a single homogeneous layer and utilized the Lucas-Kanade motion flow method to enable a high parallelism accelerator design and optimized on-chip engine computing. Furthermore, our proposed methodology achieved nearly 81% prediction accuracy with an approximately 24 FPS real-time inference throughput at 187MHz on ZCU104, which is 1.7x - 1.9x higher than the prior research. Lastly, the designed framework was benchmarked against several hardware chips for higher throughput and performance measurements and is now available as an open-source project on GitHub for training and implementation on edge platforms.

## I. INTRODUCTION

Deep Neural Networks (DNNs) with Convolutional Neural Networks (CNNs) have achieved state-of-the-art accuracy in Human Action Recognition (HAR) for understanding human interaction in video representation and time-series data [1]–[3]. Numerous methods have been proposed for DNN-based HAR, including two-stream Spatial and Temporal CNN networks [4], 3D-CNNs [5], attention mechanisms [6], and Graph Convolutional Networks (GCN) [7]. These advanced CNN architectures have progressively pushed the boundaries of what is achievable, leading to cutting-edge accuracy in recognition. The general framework for CNN-based HAR is illustrated in Fig. 1, demonstrating the learning process in DNNs network.

Nevertheless, their exceptional performance comes at the cost of a substantial number of parameters and computations involved. Numerous researchers aimed to apply compression methods to reduce HAR models' computational complexity and memory footprint, making them more suitable for edge deployment [8], [9]. For instance, they have examined direct Quantization, Quantization-Aware-Training (QAT), Pruning, and Knowledge Distillation (KD) [9]. These techniques can be used separately or combined in order to create compressed and effective models that operate in real-time performance on limited resources platforms. The selection among those methods depends on the specific requirements of the application and the trade-off between model size, performance, and available resources. For this reason, QAT Two-Stream CNNs offer the best balance of recognition performance, on-device computational/memory cost, and fewer parameters compared to 3D-CNNs, GCNs, or attention mechanisms.

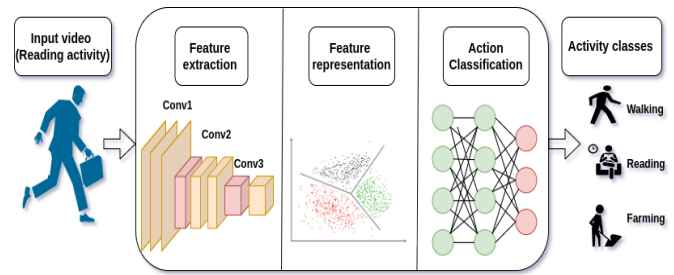

Figure 1. Human Action Recognition CNN-based framework architecture.

Several embedded platforms, such as ASIC, NVIDIA Jetson Nano, Raspberry Pi, and AMD Kria KV260, have been utilized to accelerate the two-stream CNN on the edge. Among those, the AMD System-on-Chip Field-Programmable Gate Array (SoC-FPGA) stands out as the only hardware that offers Programmable-Logic (PL) fabric coupled with an ARM-based Processing System (PS). This unique feature allows developers to reconfigure the PL fabric part and synthesize deeply pipelined custom accelerators for any algorithm. The SoC-FPGAs also allow exploring different precision optimizations to trade-off between latency, throughput, and power.

In this research project, we adopted the SimpleNet-PyTorch architecture [10] and extended it with QAT two-stream CNN. We also designed a scalable accelerator, deployed the inference on the edge, and benchmarked our design against several hardware platforms. Our key contributions are as follows:

- Parallelized the two-stream HAR SimpleNet workload by incorporating an improved fusion layer [11], consolidating all convolutional layers with batch-norm and ReLU into a single homogeneous five-layers structure [12], and utilizing Lucas-Kanade optical flow algorithm in training and implementation [13] on UCF101 [14] and UCF24 [15] datasets.

- Developed an open-source scalable and customizable inference accelerator known as (FPGA-QHAR) [16] and deployed it with PYNQ image [17] on the ZCU104 board.

- Demonstrated that the proposed methodology achieved nearly 81% accuracy and real-time throughput up to 24 FPS under 187MHz with a performance up to 120 GOP/s.

The rest of this paper is organized as follows: Section II provides background on the two-stream CNN and discusses the previous development on FPGA. Section III discusses the methodology and enhancement made to the SimpleNet-PyTorch with two-stream QHAR-based CNN architecture and the accelerator mechanism deployed on the edge. Section IV describes the experimental tools used and presents the obtained results, while Section V concludes the paper.

## II. BACKGROUND

The two-stream HAR-CNN [4] hypothesis consists of distinct networks known as spatial and temporal streams. They are used to capture both static and dynamic features respectfully present in video data. The spatial stream operates on an individual input frame (Single RGB) followed by several layers for extracting characteristics like shapes, textures, and object appearances. On the other side, the temporal stream focuses on modeling multi-frame optical flow by employing a set of input displacement vector to process the frames at time $t$ and its length $L$ subsequently into a stacked $2L$-channel to capture motions. These networks are then combined/fused using a fusion linear layer and one weighted summation to make a final prediction.

Most researchers and open-source efforts focus on utilizing 3D-CNN and GCN algorithms in order to achieve extraordinary HAR accuracy on large FPGA boards. However, these algorithms have a higher number of parameters and require more compute resources, which is inefficient for multiply and accumulate operations (MAC) and on-chip (BRAM) memory units on edge FPGAs. For this reason, the developed accelerator for two-stream SimpleNet CNN-based achieved a balanced accuracy and optimal performance for real-time inference on edge FPGAs. As a result, an efficient implementation of HAR opens the possibility of having technologies like Tesla-Autopilot [18] deployed on low-power computing platforms.

The most recent work that is in line with our idea is proposed by Lin et al. [19], which implemented an 8-bit quantized two-stream VGG7-CNN with ResNet-18 backbone on a large FPGA board (ZCU102) and achieved 12-15 FPS. Nevertheless, this research is constrained to; First, it showed one dataset result without experimenting with the network effect on smaller datasets. Second, the accelerator is not fully optimized for real-time performance, which implies that some layers are executed sequentially on the PS, causing in increased latency. Third, the model weights require larger boards and more available resources (ex. ZCU102), resulting in more power consumption. Lastly, our proposed methodology addressed these issues by balancing the network accuracy and accelerator performance to meet the needs for smaller edge devices in real-time applications.

## III. METHODOLOGY

### A. Network Architecture

To achieve optimal efficiency in both resource usage and algorithm acceleration, our customized architecture (SimpleNet) shown in Fig. 2 employs two-stream lightweight CNNs with a design process of homogeneously stacking several types of layers such as Convolutional, Batch-Normalization (BN), and Rectified Linear Unit (ReLU) in one group layer. This approach [20] allows us to easily manage the number of parameters in the network while providing better max pooling information for each semantic level. Moreover, the homogeneous layer group makes the network compression via layer fusion possible [21] to meet the constraint of edge FPGAs resources.

The CNN computations of *IFM* and *OFM* represent the input feature maps and the output feature maps, respectively, and can be mathematically expressed as shown in the equation follows:

$$OFM = \sum_{n=1}^{N} \sum_{k=1}^{k*k} IFM[n][k] \times W[n][k] + Bias \quad (1)$$

Where $N$ represents the number of *IFM* channels, $K$ is the kernel size, and $W$ is matrix weights. This operation is followed by BN to improve the training speed, accelerate the convergence, and reduce the insensitivity initialization weights of the network. It can be expressed as follows:

$$x = \gamma \frac{OFM - \mu}{\sqrt{\sigma^2 + \varepsilon}} + \beta \quad (2)$$

where $\mu$ and $\sigma^2$ represent the mean and variance, respectively, $x$ denotes the output pixel after BN, $\gamma$ refers to the scale coefficient; $\beta$ is the offset coefficient, and $\varepsilon$ indicates a very small positive number. The results are then passed through ReLU to introduces non-linearity into the network and learn complex relationships in the data, as in the equation follows:

$$y = \begin{cases} x, & x \geq 0 \\ 0.1x, & x < 0 \end{cases} \quad (3) \qquad y = \begin{cases} x, & x \geq 0 \\ 0, & x < 0 \end{cases} \quad (4)$$

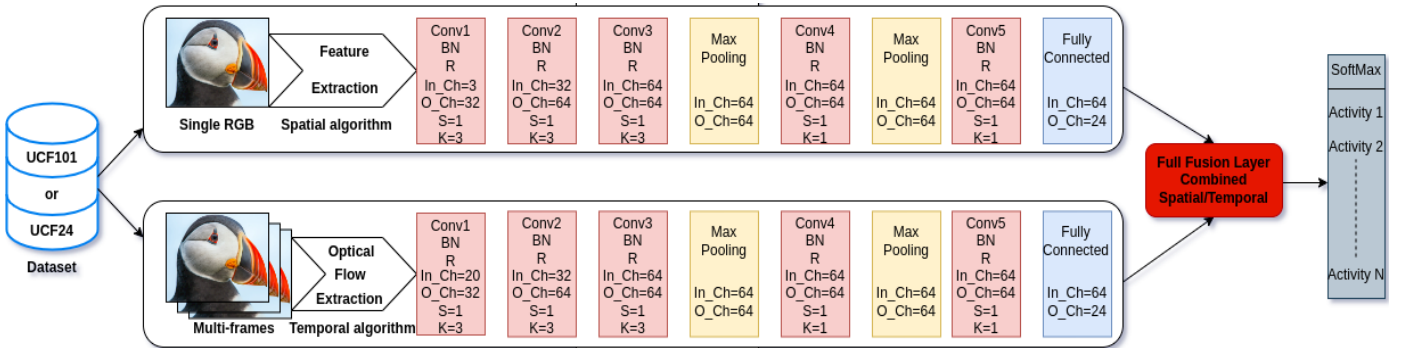

Figure 2: The proposed QAT two-stream CNN architecture combining Convolutional (Conv), Batch-Normalization (BN), and ReUL into single five layers. Input-channel (In_Ch), Output-Channel (O_Ch), Stride (S), and Kernel (K) determined the size of each layer.

Adding a pooling layer in the network after the homogeneous group helps to reduce overfitting by indicating the maximum value of the selected region. After that, the *OFM* of pooling is the *IFM* of the next homogeneous layer, which is expressed as in the following equation:

$$f = \{\max(y_1, y_2, y_3, \dots, y_n), \; \& \max pooling\} \quad (5)$$

Those operations are then followed by Fully Connected (FC) layers to perform nonlinear transformations on the features extracted by convolutional layers. In addition, the Full Fusion linear layer is introduced to minimize the computational cost of the network by combining the neurons of the spatial and temporal networks, while also enabling the learning of more complex data relationships. Subsequently, the SoftMax layer generates a probability distribution (value number *Z*) over the possible classes as indicated in the equation follows:

$$softmax(z_i) = \frac{exp(z_i)}{\sum_i exp(z_i)} \quad (6)$$

Finally, the cross-entropy loss function shown in eq. 7, is used to measure the dissimilarity between the predicted probabilities and the true labels, while the Stochastic Gradient Descent (SGD) shown in eq. 8, is computed during backpropagation to minimize the losses by updating the parameters of the network.

$$\mathcal{L}(\theta) = -\sum_{i=0}^{n} \vec{\hat{y}}_i \cdot \log(y_i) \; \Big| \; \begin{array}{l} y_i: prediction\ vector\ of\ \vec{y} \\ \hat{y}_i: ground\ truth\ label\ of\ \hat{y} \end{array} \quad (7)$$

$$\theta = \theta - \alpha \nabla_\theta J(\theta; x^{(i)}, y^{(i)}) \quad (8)$$

On the other hand, the temporal stream is computed through LK-OF proposed in [13]. This method involves the calculation of horizontal and vertical components at $t$ as a derivative of $dxt$ and $dyt$. The flow channels $dx$ and $yt$ of $L$ form consecutive frames at $2L$ input channels to reduce the processing latency. Additionally, the Lucas-Kanade method assumes that the optical flow points $v_x$ and $v_y$ are constant within a small window $W$ of size $n \times n$ pixels. Thus, the optical flow holds all pixels of the coordinates $q = (k, l)$ to window $W$ as expressed in the equation follows:

$$I_x(q)v_x + I_y(q)v_y = -I_t(q) \quad \forall q = (k, l) \in W \quad (9)$$

Where $I_x, I_y$, and $I_t$ are the partial derivative of the image intensity with respect to $v_x, v_y$. However, to obtain a compromise solution by the least squares principle, we computed a transpose matrix of $A$ as in the equation follows:

$$\begin{bmatrix} V_x \\ V_y \end{bmatrix} = \begin{bmatrix} \sum_i I_x(q_i)^2 & \sum_i I_x(q_i)I_y(q_i) \\ \sum_i I_y(q_i)I_x(q_i) & \sum_i I_y(q_i)^2 \end{bmatrix}^{-1} \begin{bmatrix} -\sum_i I_x(q_i)I_t(q_i) \\ -\sum_i I_y(q_i)I_t(q_i) \end{bmatrix} \quad (10)$$

This solves the $2 \times 2$ system, being computationally more efficient, and less sensitive to image noise than point-wise methods by assuming the flow is essentially constant in a local pixel neighborhood under consideration.

Overall, the design of both spatial and temporal model has much fewer parameters ($\approx 1.3M$) and one order of magnitude less computation overhead compared to all available alternatives [10] with a slight decrease in accuracy. Besides, the model is compressed by QAT (8-bit) and layer fusion for efficient implementation on edge SoC-FPGAs.

*B. Accelerator Design*

The proposed accelerator was built on top of Lin's project [19] and was further optimized for real-time performance. It leveraged the PL fabric to compute and parallelize the fused convolutional, batch-norm, ReLU, and max pooling layers alongside the Lukas-Kanade Optical Flow (LK-OF) separately on the processing element (PE) and the single instruction multiple data (SIMD). This approach allowed us to speed up the most intensive multiplication matrices by utilizing dedicated buffers within the memory, DSP, and LUT resources. On the other side, the remaining layers in the design (FC and Linear Fusion) were carried out on the PS part to complete the prediction action as illustrated in Fig. 3. This HW/SW scheme resulted in a reduction of the output data that is needed to be read from the DDR4 by shared buffers. It also overcomes the writing overhead latency associated with multi-dimensional arrays/frames of both spatial and temporal networks.

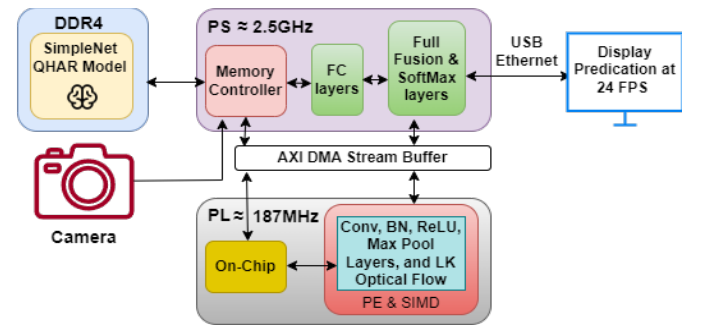

Figure 3. The mixed HW/SW accelerator for QHAR Layers on SoC-FPGA.

The developed accelerator incorporates a scheduling mechanism that efficiently transfers data between inputs, weights, and outputs using the high-performance AXI interconnect and memory controller. This is achieved through the effective execution of up-sizing, downsizing, and routing operations. The accelerator receives inputs from the processor's DDR4, executes the layers, and outputs the results back to the processor. The weights are then stored on-chip BRAM, allowing concurrent access by the partitioned PE elements in parallel, which leads to faster execution. This improves the external memory bandwidth and ensures simultaneous data read/write of *IFM*, *OFM*, and LK-OF from DDR4 to the on-chip engine and then from the on-chip engine to the compute unit. In addition, the PS part monitors the accelerator paths and can process various scalable data easily to concurrent inferences of multiple streams. Lastly, equations of the FPS performed by the accelerator and the performance measurement in Giga Operations per second (GOP/s) are illustrated respectfully:

$$Frame\ Per\ Second\ (FPS) = \frac{No.of\ Frames\ processed}{Total\ time\ (second)} \quad (11)$$

$$Performance\ GOP/s = \frac{Total\ operation}{Excuation\ Time\ (second)} \times 10^9 \quad (12)$$

*C. Loop Tiling Transformation*

Loop tiling and loop unrolling are utilized as optimization techniques for the spatial network in our architecture, specifically for the ZCU104 board. This approach improves data locality and parallelizes the design. We fetched small portions of the data on BRAM by tiling the loop alongside the row, column, and channel directions of the *IFM* and *OFM* and converted them into a one-pixel channel. Then, the corresponding data in all the input channels were element-wise multiplied by filter weights and then summed together to save the resources and reduce the number of loop iterations, resulting in more efficient instruction scheduling.

*D. Host Code*

Python with PYNQ framework [17] is used to write the two-stream application driver-code that communicates with the hardware via Linux kernel, as shown in Fig. 4. The driver loads the accelerator bitstreams and model weights into the PL part, passes input/output buffers from DDR4, and run the prediction actions inference when deployed on USB camera or video files.

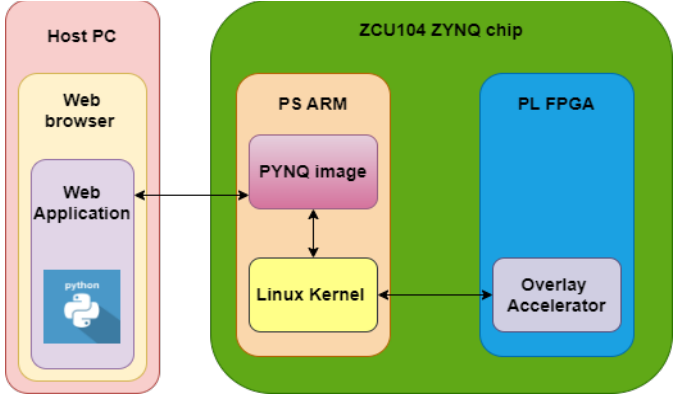

Figure 4: Software control of the accelerator via driver and operating system.

IV. EXPERIMENTAL SETUP AND RESULTS

*A. Development Environments*

The proposed two-stream QHAR SimpleNet was trained and tested on Google Colab powered by Nvidia A100 GPU with PyTorch framework and FX-Graph QAT library. Vivado High-Level Synthesis (HLS) synthesized the accelerator bitstream and analyzed the available resources on ZCU104 board after place and route. Furthermore, experiments were conducted on two well-known action recognition datasets, UCF101 and UCF24, which contain 101 and 24 action classes, accordingly, and 13, 320, and 6500 video clips respectfully.

*B. Results*

Our QHAR architecture was trained from scratch on full precision and then fine-tuned (quantized weights) to an 8-bit unsigned integer. This approach achieved ≈79% and ≈81% prediction accuracy on UCF101 and UCF24 datasets respectfully. The enhanced two-stream SimpleNet demonstrated higher accuracy results on smaller datasets with fewer classes. This is due to the fewer number of parameters (≈1.3M) and being lightweight architecture. On the other hand, the optimized accelerator operated at a frequency of 187MHz and achieved a performance of up to 120 GOP/s on ZCU104. Additionally, it was implemented with a USB camera, as shown in Fig. 5, demonstrating 22.5 FPS alongside the prediction classes. The throughput shots sometimes fluctuate in a range of (22 – 24.5) FPS when running the inference due to the model and hardware variability. We also ran the inference with video clips on ZCU104 and achieved more than 30 FPS as, shown in Fig. 6 and 7. This phenomenon is due to the camera delay in real-time, being preprocessed and post-processed on the PS part. Nevertheless, the developed accelerator-based QHAR proved to be efficient in terms of resource usage, as shown in Table 1. Lastly, our network design and accelerator inference results are reported in Table 2 showing some comparisons and benchmarks of our QHAR model (SimNet) against the CPU, GPU, NVIDIA Jetson Nano, and the previous ResNet18-FPGA study.

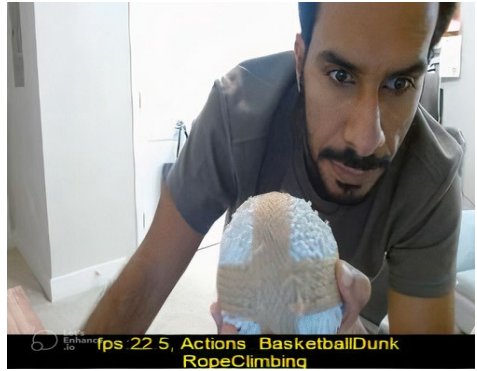

Figure 5: Action classified from USB camera.

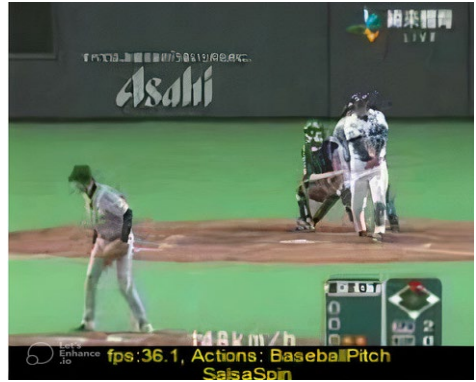

Figure 6: Action classified from video clip.

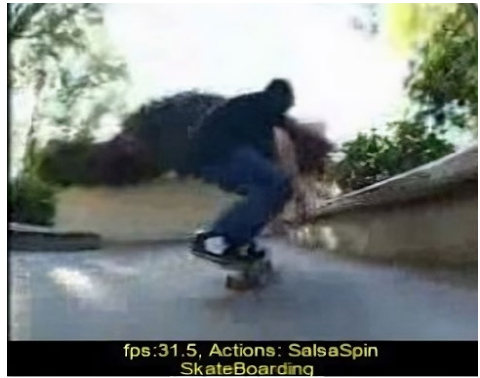

Figure 7: Action classified from video clip.

Table 1. Resource utilization comparison

| Device | ZCU104 (our) | ZCU102 (prior study [19]) |
|---|---|---|
| Flip-Flops | 83k (18%) | No report |
| LUTs | 19.1k (38%) | 227.8k (81%) |
| BRAM | 22.6k (59%) | 472 (13%) |
| DSP Slices | 967 (59%) | 1390 (54%) |
| Frequency | 187 MHz | 200 MHz |

Table 2. Benchmarking with several platforms and studies

| Platform | GPU T4 | CPU Intel Xeon ® | Jetson Nano | ZCU102 [19] | ZCU104 (our) |
|---|---|---|---|---|---|
| Model | SimNet | SimNet | SimNet | ResNet18 | SimNet |
| Dataset | UCF24 | UCF24 | UCF24 | UCF101 | UCF24 |
| Bit width | 8-bit | 8-bit | 8-bit | 8-bit | 8-bit |
| M-size | 5.1 MB | 5.1 MB | 5.1 MB | 22.3 MB | 5.1 MB |
| Accuracy | 81% | 81% | 81% | 86% | 81% |
| Frequency | 585MHz | 2.2 GHz | 1.5GHz | 200MHz | 187MHz |
| GOP/s | - | - | - | 4.12 | 120 |
| FPS | ≈51 | ≈1 | ≈9 | ≈15 | ≈24 |

## V. CONCLUSION AND FUTURE WORK

This research project introduced a scalable real-time QHAR-based hardware accelerator for two-stream CNN on SoC-FPGA. The proposed technique optimized SimpleNet with homogeneous layers and LK-OF estimation method for less computation and higher accuracy. As a result, a range of 1.7x - 1.9x higher throughput was achieved along with fewer network parameters compared to the previous research [19]. This optimistic end-to-end open-source framework [16] can be more customized to work with different boards and datasets. Future work includes improving our architecture accuracy and developing an open-source multimodal real-time accelerator for Advanced Driver Assistance System with HAR that can recognize not only driver actions but also actions of pedestrians, cyclists, and other vehicles on the road, similar to Tesla Autopilot.